# Clear the Fog: Combat Value Assessment in Incomplete Information Games with Convolutional Encoder-Decoders

**Hyungu Kahng[2], Yonghyun Jeong[1], Yoon Sang Cho[2], Gonie Ahn[2], Young Joon Park[2], Uk Jo[1], Hankyu Lee[2], Hyungrok Do[2], Junseung Lee[1], Hyunjin Choi[1], Iljoo Yoon[1], Hyunjae Lee[1], Daehun Jun[1], Changhyeon Bae[1], Seoung Bum Kim[2]**

Samsung SDS[1], Korea University[2]
hgkahng@korea.ac.kr, yonghyunjjj@gmail.com, yscho187.5@gmail.com, gonie32@gmail.com, yjpark1@korea.ac.kr,
verystrongjoe@gmail.com, hankyulee86@korea.ac.kr, hyungrok.do11@gmail.com, ppsd090@gmail.com, anneshj@naver.com,
vpds@naver.com, oglee815@gmail.com, gangoku@naver.com, charlie.bae@samsung.com, sbkim1@korea.ac.k

## Abstract

StarCraft, one of the most popular real-time strategy games, is a compelling environment for artificial intelligence research for both micro-level unit control and macro-level strategic decision making. In this study, we address an eminent problem concerning macro-level strategic decision making, known as the "fog-of-war", which rises naturally from the fact that information regarding the opponent's state is always provided in incomplete form. For intelligent agents to play like human players, it is obvious that making accurate predictions of the opponent's status under incomplete information will increase its chance of winning. To reflect this fact, we propose a convolutional encoder-decoder architecture that predicts potential counts and locations of the opponent's units based on only partially visible, noisy information. To evaluate the performance of our proposed method, we train an additional classifier on the encoder-decoder output to predict the game outcome (win or lose). Finally, we designed an agent incorporating the proposed method and conducted simulation games against rule-based agents to demonstrate both effectiveness and practicality. All experiments were conducted on actual game replay data acquired from professional players.

## Introduction

The common goal in real-time strategy (RTS) games is to gather resources efficiently, build adversarial strategies, produce units, effectively combat and finally eliminate all opponent units to seize victory. However, it is more than challenging to design human-level artificial intelligence (AI) players, especially because RTS games involve complex dynamics with a state and action space larger than other classic board games (Shantia A., 2011). This is why RTS games such as StarCraft have aroused continuous attention from the AI research community in recent years. In particular, both human and AI players in RTS games face the problem of incomplete information, called "fog-of-war" (Ontan'on et al. 2013), in which players are only exposed to regions in the map where friendly units reside. For this reason, human players attempt to acquire opponent's information as much as possible by sending scout units to unseen areas. Based on this information, human players by intuition estimate the total counts and positions of opponent units. Removing situational uncertainties due to 'fog-of-war' by making accurate estimates of the counts and positions of the opponent's units earns the player an overwhelming advantage over the opponent, and is highly likely to win the match. (Cho H., 2016). In fact, it is well-known that a high-level human player never loses when the game is played unfairly under complete information using map hacks. It is obvious to say that with complete information of the game state, opponent's strategies including build orders and attacks can be easily predicted in advance.

This study proposes a method based on convolutional encoder-decoders to predict the opponent's hidden unit information covered by the fog-of-war nature of the game. To the best of our knowledge, this paper is the first to introduce convolutional encoder-decoder networks as a means of directly tackling the fog-of-war problem in StarCraft. Primarily, we provide visualizations of the predictions to demonstrate its predictive performance. We further extend our work and evaluate its performance by training a CNN classifier to predict the game outcomes at an intermediate stage of the game. Finally, we deploy it on a AI agent capable of utilizing the predictions to determine the most effective combat time real-time. Simulations are conducted to demonstrate the effectiveness and practicality of our approach.

## Related Work

**StarCraft AI**

Real-time strategy (RTS) games propose compelling challenges to the artificial intelligence (AI) research community. In particular, the game of StarCraft is an environment of complex dynamics which accommodates many intriguing features for research, making it an appropriate testbed for human-level AI. First of all, an infinite amount of training data can be obtained to train deep neural networks, which has recently positioned itself as a standard in machine learning-based AI (Vinyals, O. et al., 2017). In addition, after its first release in 1998, StarCraft has been the most widely played RTS game, which means that a sufficient pool of game replays are easily available. Similarly, additional data gathering is cost-effective in the sense that data can be easily generated from continuous game simulations. Another rich feature is that deterministic outcomes are provided for every game played, which can be utilized as class labels in supervised learning or as rewards in reinforcement learning, to name just a few.

The StarCraft AI research community often divides tasks into two different categories: micro-level and macro-level decision making (Ontanón, S. et al., 2013). As the name implies, micro-level decision making refers to the ability to reactively control individual units to perform unit-level actions such as moving, targeting and firing. Research on such micromanagement tasks are usually conducted on a simplified mini-game exclusively designed for their specific goals to reducing the state and action space of the task (Wender S. et al., 2012). For the moment, most multi-agent reinforcement learning research fall under this category (Usunier N. et al., 2017, Wender S. et al., 2016). On the other hand, macro-level decision making refers to long-term strategy planning, ranging from spatial reasoning of building placement or base expansion to temporal reasoning of appropriate unit production and expansion times, resource gathering, and timing attacks and retreats (Weber B. et al., 2011). Using game replays, Bayesian models (Synnaeve G. et al., 2011) were used for strategy prediction, while Justesen N. et al (2017) made a neural network-based approach to learn a data-driven policy for strategy construction. Recent work also includes strategy selection from a predefined set of strategies, by formulating a metagame and estimating its Nash Equilibrium (Tavares A. et al., 2016), and game state evaluation based on deep convolutional neural networks (Stanescu M. et al., 2016).

**Challenges**

A major challenge that distinguishes StarCraft from traditional board games lies in the computational complexity due to the game's complicated dynamics. State-of-the-art AIs commonly suffer from the enormous state and action spaces of the game which naturally rises from the intractable combinations of units and real-time decisions made simultaneously. As shown in the [Table 1], the complexity of StarCraft is many orders of magnitude larger than other traditional games that have been proved to be conquered (Ontanón, S. et al., 2013).

|  | Chess | Porker | Go | StarCraft |
|---|---|---|---|---|
| Complexity | $10^{50}$ | $10^{80}$ | $10^{170}$ | $10^{1685}$ |

*[Table 1] Computational complexity of games*

Alongside with the complexity issue, AI algorithms for StarCraft must cope with incomplete information regarding the opponent's base, namely the 'fog-of-war'. In other words, players' view of the game is restricted to only those regions occupied by friendly units. Partial knowledge of the opponent's strategy can be obtained by scouting, which refers to the task of sending units to invisible terrains of the map. However, it is often the case that scout units do not survive long enough to provide continuous and complete information of the opponent's tactics. While experienced human players somehow easily manage to deduce opponent's strategy under limited information, it remains a burden for StarCraft AIs in the context of macro-level decision making. To the best of our knowledge, previous StarCraft AI research have not yet tackled the 'fog-of-war' situation directly (Erikson G. et al., 2014, Stanescu M. et al., 2016). In other words, despite its importance, either complete information was assumed or the information lying under the fog was ignored.

**Convolutional Neural Networks**

In recent years, deep convolutional neural networks (CNN) have become a standard machine learning architecture in the field of computer vision and natural language processing, to achieve remarkable performance in a variety of tasks such as image classification, video prediction, object detection, instance segmentation, text classification and machine translation, At the core of its accomplishments lies its unique structural capability of learning hierarchical representations from the input data which are rich in the context of minimizing a task-specific loss function. A stack of convolution layers with localized filters builds higher-level abstractions from low-level features such as pixels, characters and words.

Numerous research domains are taking advantage of the spatially invariant property of deep CNNs, and the field of game AI research is no exception. Mnih V. et al (2015) demonstrated a CNN agent trained with deep reinforcement learning that surpasses human performance in playing Atari video games. AlphaGo, the first AI agent to defeat a human professional Go player, consists of two deep CNNs that recognize the 19×19 board to predict the value of the state and the best moves (Silver D. et al., 2016). In the context of StarCraft AI, Vinyals O. et al (2017) and Zambaldi V. et al (2018) trained CNNs on raw game frames to play minigames at a human level. These achievements serve as a basis for extending applications of CNN architectures to tackle problems in StarCraft in the context of incomplete information.

# Method

In this paper, we propose a convolutional encoder-decoder network sophisticatedly designed to retrieve information covered by fog in StarCraft games. Provided incomplete information of the current state, the proposed architecture learns to predict both potential locations and counts of the enemy's units in an end-to-end fashion. We take three different approaches to demonstrate the performance of our proposed method in recovering hidden information: 1) visualization, 2) classification, and 3) simulation. Details on datasets, network specifications and evaluation measures will be provided throughout the rest of the paper.

**Dataset**

We received 2,200 actual replay files from 11 professional StarCraft players. We limit the games to those between the Terran and Protoss army to simplify the framework. Using the Brood War Application Programming Interface (BWAPI), a free and open source tool provided for StarCraft AI research, we parsed the replay files and saved frame-level game logs every three seconds. We also performed domain-specific data cleaning in order to eliminate uninformative frames in the too-early and too-late stages of the game, where it is either impractical or meaningless for the purpose of our analysis. Our final dataset consists of 500,000 game frames.

As previous research shows, not only the counts but also the positions of units are important features of the game (Stanescu M. et al., 2016). To preserve the spatial information as much as possible, game states are expressed as feature maps. The Terran and Protoss army have 34 and 32 distinguishable units respectively, thus we chose to assign a unique feature map for each unit type. In addition, since the original 4,096 × 4,096 pixel grid is too large to process, we downsized both height and width to 32. As depicted in [Figure 1], a single game frame of 4096×4096×3 pixels is converted into a 32×32×66 feature map.

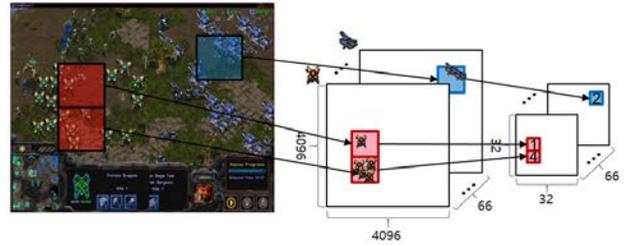

*[Figure 1] Raw game frames to feature maps*

The final dataset for training our proposed model consists of (X, Y) pairs, where X denotes the noisy feature maps resulting from fog-of-war, and Y denotes the clean feature maps with complete information.

**Convolutional Encoder-Decoder Architecture**

The proposed network consists of a convolutional encoder and a decoder network. The encoder receives a noisy 32 × 32 × 66 feature map of the game state and performs a cascade of convolution operations to extract meaningful features representing the input. The decoder takes this high-level abstraction and performs transpose convolution operations to retrieve the complete data without fog. The model is trained to minimize mean squared error between the input and output. Following the style of the VGG network (Simonyan K. et al., 2014), filters sizes are fixed to 3×3, while the number of filters are doubled when the feature map size is halved. The network is fully convolutional, without any spatial pooling or fully-connected layers. Instead of pooling, we halve the feature maps' height and width by using a convolutional stride of 2. The removal of fully-connected layer in bottlenecks is intended to preserve spatial information from input to output. Every convolution layer is followed by a Rectified Linear Unit (ReLU) to perform nonlinear transformation (Nair, V. et al., 2010). Another feature of the architecture is symmetricity, meaning that each convolution layer in the encoder is paired with a transpose convolution layer in the decoder to share its trainable weights. Weight sharing acts as a regularization leading to faster convergence and prevents the model from learning degenerate solutions (Masci J. et al., 2011).

Regarding our task of retrieving hidden information covered by fog-of-war, a direct skip connection (He K. et al., 2015) from input to output are used to alleviate the trivial task of having to predict units already visible in the input. We found this choice of skip connection do not only accelerate training but also enhance the quality of retrieved outputs. The whole architecture is characterized in [Figure 2].

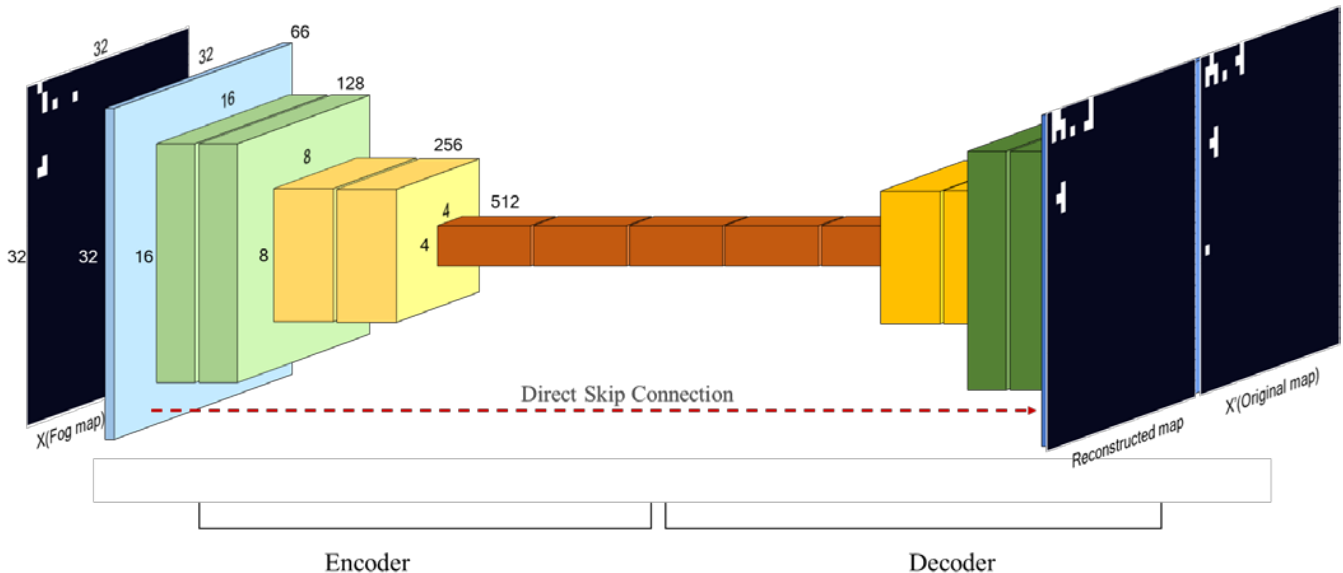

*[Figure 2] Architecture of the proposed convolutional encoder-decoder network*

## Experiments

**Network Training**

From the dataset of 500,000 game frames, we use 450,000 (90%) frames to train the convolutional encoder-decoder, and 50,000 (10%) frames to validate its performance. We made sure that frames from the same replay fall into only one of train or validation sets to ensure temporal independency between the two sets. If not, the validation set will be inappropriate as a measure of the model's generalization ability.

We run 500 epochs of training with a batch size of 512, and chose the epoch with the smallest validation loss as our best model. Stochastic gradient descent was performed with the Adam optimizer with a fixed learning rate 0.001, which guarantees fast convergence (Kingma D. et al., 2014). All trainable network parameters were initialized with Xavier initialization (Glorot X. et al., 2010). All experiments were conducted on a single GTX 1080 Ti, with Tensorflow (Abadi M. et al., 2016).

**Evaluation & Results**

(Visualization) Concrete examples comparing the decoder predictions with the noisy data and the original complete data is visualized as heatmaps in [Figure 3]. The feature maps are downsized by summation to a 8×8 grid for simplicity. In each row of the figure, the first plot depicts the noisy corrupted input, the second row depicts the original complete data, and the last plot is the prediction made by the decoder. Even though the predictions are not pointwise accurate, the results imply some interesting properties. The first row contains heatmaps of the Protoss unit 'Probe'. In most of the cases, probes are accurately predicted. The second row contains heatmaps of the combat unit 'Dragoon'. This is important because predicting positions of attack units is crucial to estimating combat time. The third row contains heatmaps for the unit 'Pylon'. Here, two extra expansions are inferred, which can inform the Terran player the opponent is playing defensively.

(Classification) We take another indirect approach to evaluate the quality of outputs generated by the proposed convolutional encoder-decoder architecture. We train three additional CNN classifiers to predict the winner of the game, on the 1) noisy incomplete data, 2) original complete data, and 3) retrieved decoder predictions respectively. All three models use the same architecture: five convolution layers with 3x3 filters and two max pooling layers followed by ReLU. The winner prediction model takes a $32 \times 32 \times 66$ feature map as input, and predicts the actual winner of the game by softmax classification.

|  | Noisy incomplete data | Clean complete data | Retrieved decoder prediction |
|---|---|---|---|
| Test Acc. | 65.1% | 87.5% | 82.5% |
| F1-score | 53.3% | 87.1% | 81.1% |

*[Table 2] Test accuracy of CNN classifer trained on three different types of inputs*

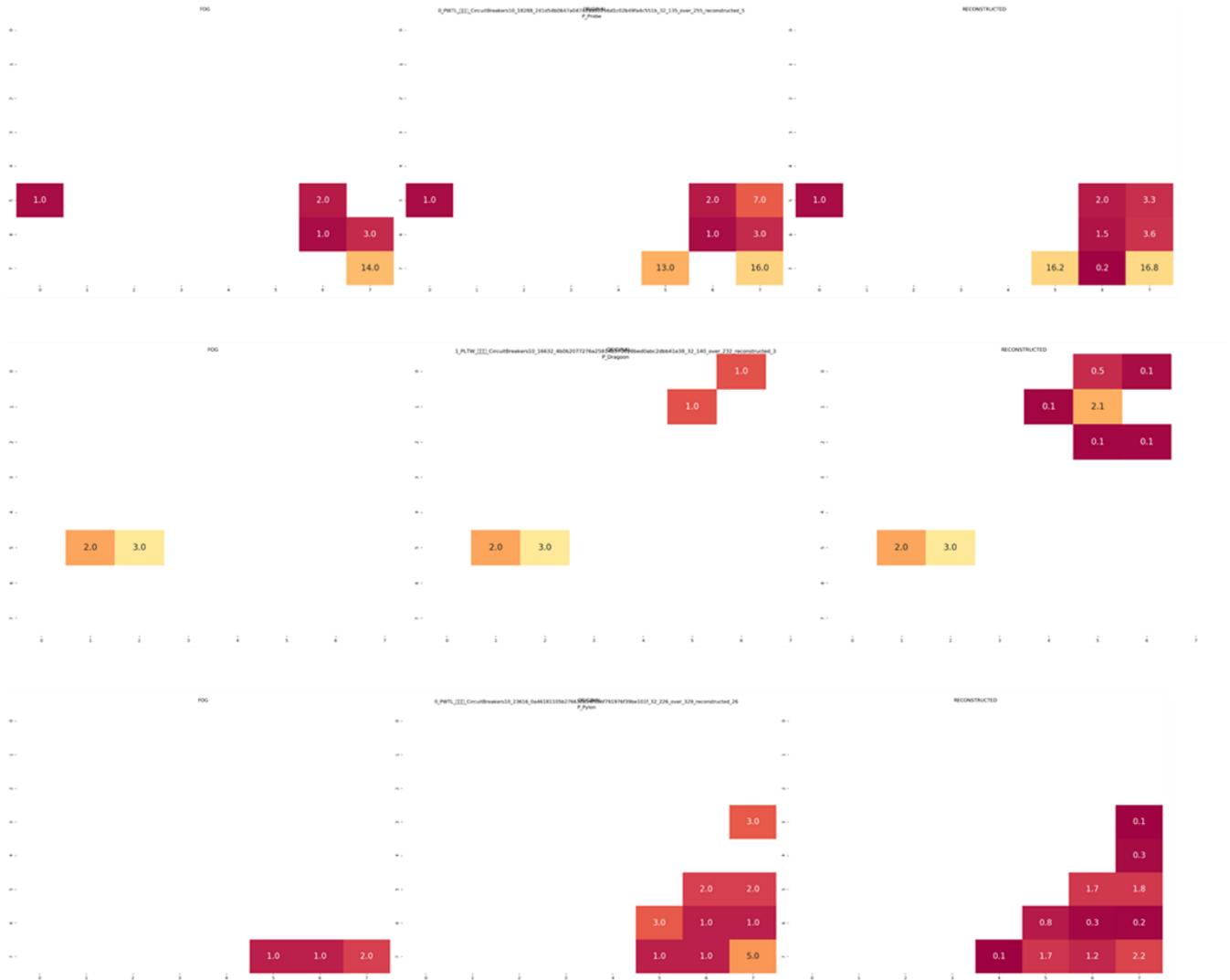

*[Figure 3] Examples of prediction made by the proposed convolutional encoder-decoder*

We evaluate the performance of all three classifiers on a test set of 20 game frames considered to be the critical point that determines victory or defeat of the game. The class labels are balanced: Terran wins 10 games, and Protoss wins 10 games. As shown in the [Table 2], the classifier trained on the decoder predictions achieve a test accuracy of 82.5%, which is by a considerable margin larger than 65.1%, the test accuracy of the classifier trained on noisy incomplete data. Intuitively, training on clean complete data should deservedly obtain the highest accuracy (87.5%). From the results, we can infer that the predictions made by the proposed encoder-decoder model are semantically rich and similar to the original complete data.

(Simulation) To evaluate the effectiveness and practicality of the proposed method, we mount the convolutional encoder-decoder and the trained CNN classifier on a rule-based StarCraft AI agent to aid decision making of the best combat times, and examine its performance gain over plain rule-based agents. Both agents play Terran, and is based on the open-source uAlbertaBot. The plain rule-based agent determines the combat time by the ratio of opponent's units to the number of friendly units. Since the opponent information is incomplete, we multiply a hand-crafted correction coefficient to adjust the ratio. In contrast, our proposed agent determines the combat time based on the winning probability predicted by the trained CNN classifier. In this test, the probability threshold was fixed to 69%, meaning that if the CNN classifier predicts that there is a winning probability higher than 69%, the agent goes to combat given that the level of Terran's Vehicle Weapons is upgraded.

On our simulation environment, we run 10 full time games against each of the eight open-source Protoss agents: Wuli, SkyNet, TomasVajda, Andrew, MegaBot, PurpleWave, Xelnaga, and Locutus. As shown in [Figure 4], the proposed

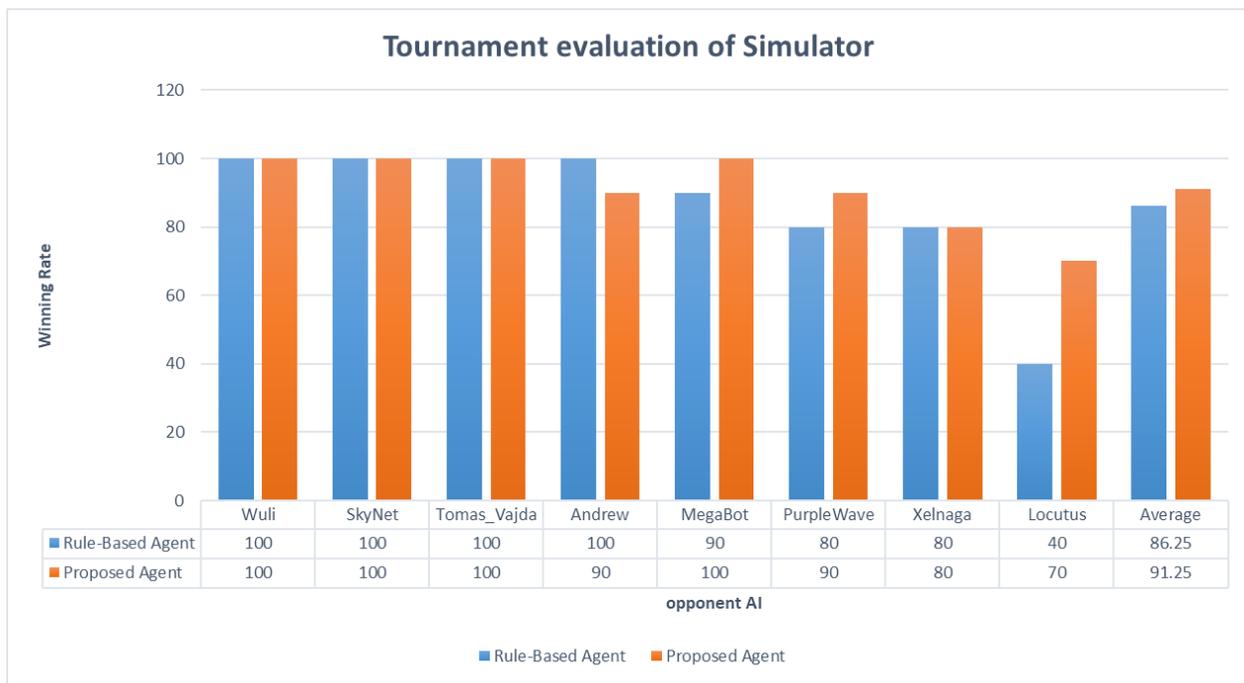

*[Figure 4] Comparison of simulation results of rule-based agents and our AI agent*

agent shows higher winning rates than the plain rule based agent against seven out of eight opponents.

## Conclusions

In this paper, we proposed a convolutional encoder-decoder network for StarCraft AI agents capable of predicting the potential counts and positions of opponent's units. Through three different evaluation measures we have shown that the proposed method is effective in the tackling fog-of-war, and at the same time practical to be implemented in a StarCraft AI agent. The CNN-based winner predictor shows superior performance when trained on decoder predictions than when trained on noisy incomplete inputs. Simulation results show that agents incorporating the proposed methods outperform a basic rule-based agent in seven out of eight games against popular AI agents.

Clearing the fog in incomplete information games is a fundamental yet relatively unexplored topic of research in the context of macro-level decision making. We hope to improve our convolutional encoder-decoder with more advanced techniques that have been introduced in the literature of computer vision.

## References

Abadi, M., Barham, P., Chen, J., Chen, Z., Davis, A., Dean, J., ... & Kudlur, M. (2016, November). Tensorflow: a system for large-scale machine learning. In OSDI (Vol. 16, pp. 265-283).

Buro, M., & Churchill, D. 2012. Real-time strategy game competitions. *AI Magazine*, 33(3), 106.

Cho, H., Park, H., Kim, C. Y., & Kim, K. J. 2016. Investigation of the Effect of "Fog of War" in the Prediction of StarCraft Strategy Using Machine Learning. *Computers in Entertainment (CIE)*, 14(1), 2.

Dereszynski, E. W., Hostetler, J., Fern, A., Dietterich, T. G., Hoang, T. T., & Udarbe, M. 2011. Learning Probabilistic Behavior Models in Real-Time Strategy Games. In *AIIDE*.

Erickson, G. K. S., & Buro, M. 2014. Global State Evaluation in StarCraft. In *AIIDE*.

Glorot, X., & Bengio, Y. (2010, March). Understanding the difficulty of training deep feedforward neural networks. In Proceedings of the thirteenth international conference on artificial intelligence and statistics (pp. 249-256).

He, K., Zhang, X., Ren, S., & Sun, J. (2016). Deep residual learning for image recognition. In Proceedings of the IEEE conference on computer vision and pattern recognition (pp. 770-778).

Justesen, N., & Risi, S. 2017. Learning macromanagement in starcraft from replays using deep learning. *arXiv preprint arXiv*:1707.03743.

Kingma, D. P., & Ba, J. (2014). Adam: A method for stochastic optimization. arXiv preprint arXiv:1412.6980.

Masci, J., Meier, U., Cireşan, D., & Schmidhuber, J. (2011, June). Stacked convolutional auto-encoders for hierarchical feature extraction. In International Conference on Artificial Neural Networks (pp. 52-59). Springer, Berlin, Heidelberg.

Mnih, V., Kavukcuoglu, K., Silver, D., Rusu, A. A., Veness, J., Bellemare, M. G., ... & Petersen, S. (2015). Human-level control through deep reinforcement learning. Nature, 518(7540), 529.

Nair, V., & Hinton, G. E. (2010). Rectified linear units improve restricted boltzmann machines. In Proceedings of the 27th international conference on machine learning (ICML-10) (pp. 807-814).

Ontanón, S., Synnaeve, G., Uriarte, A., Richoux, F., Churchill, D., & Preuss, M. 2013. A survey of real-time strategy game AI research and competition in StarCraft. *IEEE Transactions on Computational Intelligence and AI in games*, 5(4), 293-311.

Peng, P., Yuan, Q., Wen, Y., Yang, Y., Tang, Z., Long, H., & Wang, J. 2017. Multiagent bidirectionally-coordinated nets for learning to play starcraft combat games. *arXiv preprint arXiv*:1703.10069.

Robertson, G., & Watson, I. 2015. Building behavior trees from observations in real-time strategy games. In *Innovations in Intelligent SysTems and Applications (INISTA), 2015 International Symposium on* (pp. 1-7). IEEE.

Shantia, A., Begue, E., & Wiering, M. 2011. Connectionist reinforcement learning for intelligent unit micro management in starcraft. In *Neural Networks (IJCNN), The 2011 International Joint Conference on* (pp. 1794-1801). IEEE.

Silver, D., Huang, A., Maddison, C. J., Guez, A., Sifre, L., Van Den Driessche, G., ... & Dieleman, S. (2016). Mastering the game of Go with deep neural networks and tree search. nature, 529(7587), 484.

Simonyan, K., & Zisserman, A. (2014). Very deep convolutional networks for large-scale image recognition. arXiv preprint arXiv:1409.1556.

Stanescu, M., Barriga, N. A., Hess, A., & Buro, M. 2016. Evaluating real-time strategy game states using convolutional neural networks. In *Computational Intelligence and Games (CIG), 2016 IEEE Confrence on* (pp. 1-7). IEEE.

Synnaeve, G., & Bessiere, P. 2011. A Bayesian Model for Plan Recognition in RTS Games Applied to StarCraft. In *AIIDE*.

Tavares, A., Azpurua, H., Santos, A., & Chaimowicz, L. 2016. Rock, paper, starcraft: Strategy selection in real-time strategy games. In *The Twelfth AAAI Conference on Artificial Intelligence and Interactive Digital Entertainment (AIIDE-16)*.

Usunier, N., Synnaeve, G., Lin, Z., & Chintala, S. 2016. Episodic exploration for deep deterministic policies: An application to starcraft micromanagement tasks. *arXiv preprint arXiv*:1609.02993.

Vinyals, O., Ewalds, T., Bartunov, S., Georgiev, P., Vezhnevets, A. S., Yeo, M., ... & Quan, J. (2017). Starcraft ii: A new challenge for reinforcement learning. arXiv preprint arXiv:1708.04782.

Wender, S., & Watson, I. 2012. Applying reinforcement learning to small scale combat in the real-time strategy game StarCraft: Broodwar. In *Computational Intelligence and Games (CIG), 2012 IEEE Conference on* (pp. 402-408). IEEE.

Wender, S., & Watson, I. 2016. Combining case-based reasoning and reinforcement learning for tactical unit selection in real-time strategy game AI. In *International Conference on Case-Based Reasoning* (pp. 413-429). Springer, Cham.

Zambaldi, V., Raposo, D., Santoro, A., Bapst, V., Li, Y., Babuschkin, I., ... & Shanahan, M. (2018). Relational Deep Reinforcement Learning. arXiv preprint arXiv:1806.01830.